%% file: main.tex
\documentclass[%
 reprint,
 amsmath,
 amssymb,
 aps,
]{revtex4-1}

\usepackage[utf8]{inputenc}
\usepackage{subfiles}
\usepackage{graphicx}
\usepackage{hyperref}
\usepackage[caption=false]{subfig}
\usepackage{amssymb}
\usepackage{amsmath}
\usepackage{braket}
\usepackage{mathrsfs}
\usepackage{amsbsy}
\usepackage{physics}
\usepackage{color, soul}
\usepackage[nonumberlist, acronym, toc]{glossaries}
\glsdisablehyper
\usepackage{algorithm}
\usepackage[noend]{algpseudocode}
\usepackage{enumitem}

\usepackage{multirow} 

\usepackage{bm}

\DeclareMathOperator{\EX}{\mathbb{E}}
\mathchardef\mhyphen="2D

\makeatletter
     \renewcommand*\l@figure{\@dottedtocline{1}{1em}{3.2em}}
     \renewcommand*\l@table{\@dottedtocline{1}{1em}{3.2em}}     
\makeatother

\makeglossaries
\loadglsentries{glossary.tex}

\begin{document}


\title{Quantum-assisted associative adversarial network: Applying quantum annealing in deep learning}

\author{Max Wilson}
\affiliation{Quantum Artificial Intelligence Lab., NASA Ames Research Center, Moffett Field, CA 94035, USA}
\affiliation{Quantum Engineering CDT, Bristol University, Bristol, BS8 1TH, UK}

\author{Thomas Vandal}
\affiliation{NASA Ames Research Center / Bay Area Environmental Research Institute, Moffett Field, CA 94035, USA}

\author{Tad Hogg}
\affiliation{Quantum Artificial Intelligence Lab., NASA Ames Research Center, Moffett Field, CA 94035, USA}

\author{Eleanor Rieffel}
\affiliation{Quantum Artificial Intelligence Lab., NASA Ames Research Center, Moffett Field, CA 94035, USA}

\date{\today}

\begin{abstract}
We present an algorithm for learning a latent variable generative model via generative adversarial learning where the canonical uniform noise input is replaced by samples from a graphical model. This graphical model is learned by a Boltzmann machine which learns low-dimensional feature representation of data extracted by the discriminator. A quantum annealer, the D-Wave 2000Q, is used to sample from this model. This algorithm joins a growing family of algorithms that use a quantum annealing subroutine in deep learning, and provides a framework to test the advantages of quantum-assisted learning in GANs. Fully connected, symmetric bipartite and Chimera graph topologies are compared on a reduced stochastically binarized MNIST dataset, for both classical and quantum annealing sampling methods. The quantum-assisted associative adversarial network successfully learns a generative model of the MNIST dataset for all topologies, and is also applied to the LSUN dataset bedrooms class for the Chimera topology. Evaluated using the Fr\'{e}chet inception distance and inception score, the quantum and classical versions of the algorithm are found to have equivalent performance for learning an implicit generative model of the MNIST dataset. 
\end{abstract}

\maketitle

\input{layout.tex}

\bibliography{bibliography}
\bibliographystyle{ieeetr}

\end{document}

%% file: glossary.tex
\newacronym{ml}{ML}{Machine Learning}
\newacronym{es}{ES}{Earth Sciences}
\newacronym{rbm}{RBM}{Restricted Boltzmann Machine}
\newacronym{cd}{CD}{Contrastive Divergence}
\newacronym{dbn}{DBN}{Deep Belief Net}
\newacronym{qa}{QA}{Quantum Annealing}
\newacronym{qaaan}{QAAAN}{Quantum Assisted Associative Adversarial Network}
\newacronym{mnist}{MNIST}{Modified National Institute of Standards and Technology database}
\newacronym{cdt}{CDT}{Centre for Doctoral Training}
\newacronym{gan}{GAN}{Generative Adversarial Network}
\newacronym{sgd}{SGD}{Stochastic Gradient Descent}
\newacronym{qip}{QIP}{Quantum Information Processing}
\newacronym{qc}{QC}{Quantum Computing}
\newacronym{elbo}{ELBO}{Evidence Lower Bound}
\newacronym{vae}{VAE}{Variational Autoencoder}
\newacronym{qvae}{QVAE}{Quantum Variational Autoencoder}
\newacronym{snail}{SNAIL}{Small Noisy Annealing Implementation Learning}
\newacronym{bm}{BM}{Boltzmann Machine}
\newacronym{kl}{KL}{Kullback-Leibler}
\newacronym{fid}{FID}{Frech\'et Inception Distance}
\newacronym{is}{IS}{Inception Score}

%% file: layout.tex
\section{Introduction}\label{s:intro}
\input{sections/introduction}

\section{Background}\label{s:background}
\input{sections/background}


\section{Quantum-assisted associative adversarial network}\label{s:methods_a}
\input{sections/methods.tex}

\section{Results \& Discussion}\label{s:discussion}
\input{sections/discussion.tex}

\section{Conclusions}\label{s:conclusions}
\input{sections/conclusions.tex}



%% file: sections/introduction.tex
The ability to efficiently and accurately model a dataset, even without full knowledge of why a model is the way it is, is a valuable tool for understanding complex systems. \gls{ml}, the field of data analysis algorithms that create models of data, is experiencing a renaissance due to the availability of data, increased computational resources and algorithm innovations, notably in deep neural networks \cite{royalsoc2017ml, silver2016mastering}. Of particular interest are unsupervised algorithms that train generative models. These models are useful because they can be used to generate new examples representative of a dataset. 

A \gls{gan} is an algorithm which trains a latent variable generative model with a range of applications including image or signal synthesis, classification and image resolution. The algorithm has been demonstrated in a range of architectures, now well over 300 types and applications, from the \gls{gan} zoo \cite{isola2017image,radford2015unsupervised, ledig2016photo}. Two problems in \gls{gan} learning are non-convergence, oscillating and unstable parameters in the model, and mode collapse, where the generator only provides a small variety of possible samples. These problems have been addressed previously in existing work including energy based \gls{gan}s \cite{zhao2016energy} and the Wasserstein \gls{gan} \cite{arjovsky2017wasserstein, gulrajani2017improved}. Another proposed solution involves replacing the canonical uniform noise prior of a \gls{gan} with a prior distribution modelling low-dimensional feature representation of the dataset. Using this informed prior may alleviate the learning task of the generative network, decrease mode-collapse and encourage convergence \cite{arici2016associative}.




This feature distribution is a rich and low-dimensional representation of the dataset extracted by the discriminator in a \gls{gan}. A generative probabilistic graphical model can learn this feature distribution. However, given the intractability of calculating the exact distribution of the model, classical techniques often use approximate methods for sampling from restricted topologies, such as contrastive divergence, to train and sample from these models. Quantum annealing, a quantum optimisation algorithm, has been shown to sample from a Boltzmann-like distribution on near-term hardware \cite{benedetti2017quantum, amin2016quantum}, which can be used in the training of these types of models. In the future, quantum annealing may decrease the cost of this training by decreasing the computation time \cite{biamonte2017quantum}, energy usage \cite{ciliberto2018quantum}, or improve performance as quantum models \cite{kappen2018learning} may better represent some datasets.

\begin{figure}[h]
    \centering
    \includegraphics[width=0.50\textwidth]{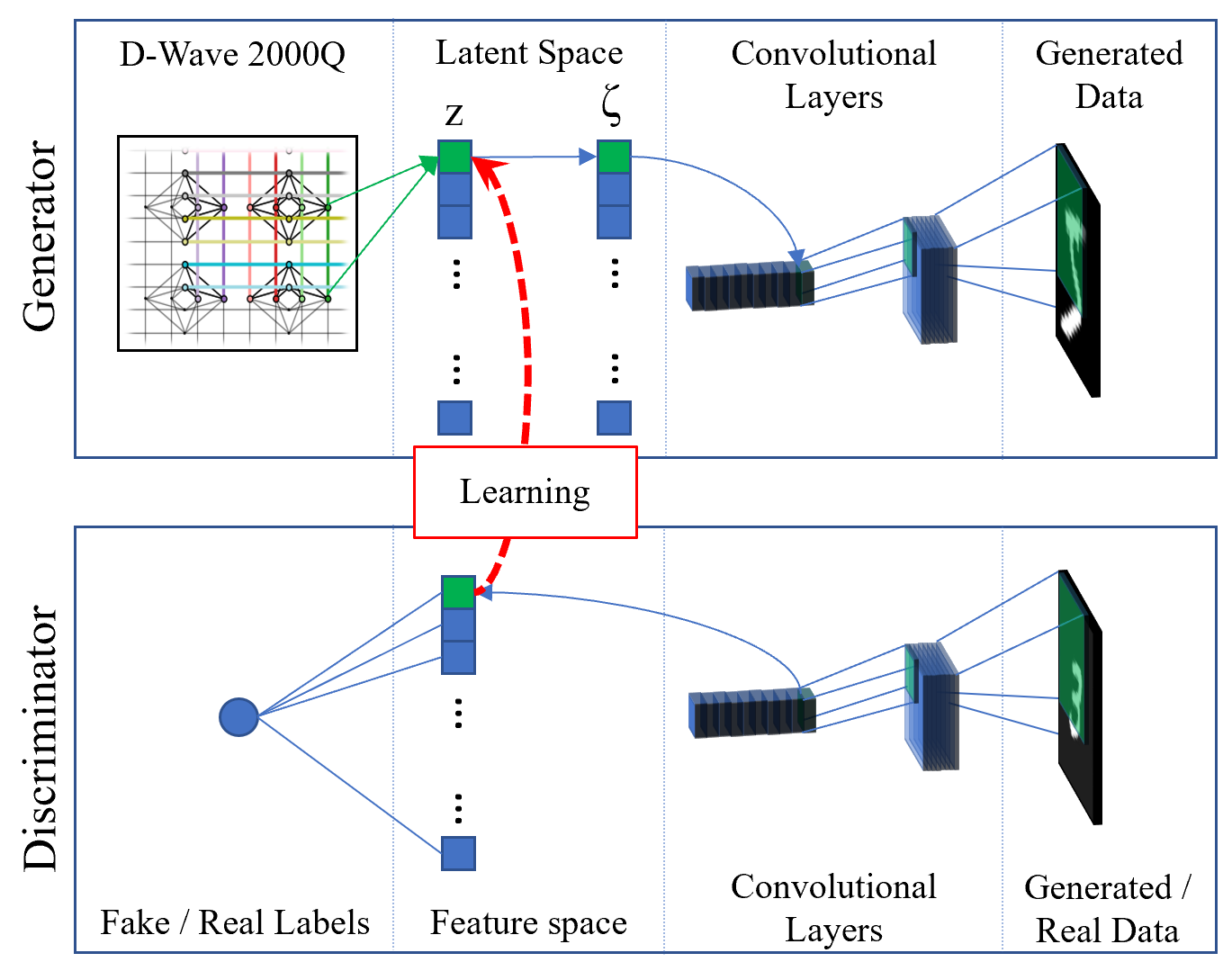}
    \caption{The inputs to the generator network are samples from a Boltzmann distribution. A \gls{bm} trains a model of the feature space in the generator network, indicated by the Learning. Samples from the quantum annealer, the D-Wave 2000Q, are used in the training process for the \gls{bm}, and replace the canonical uniform noise input to the generator network. These discrete variables $\mathbf{z}$ are reparametrised to continuous variables $\boldsymbol{\zeta}$ before being processed by transposed convolutional layers. Generated and real data are passed into the convolutional layers of the discriminator which extracts a low-dimensional representation of the data. The \gls{bm} learns a model of this representation. An example flow of information through the network is highlighted in green. In the classical version of this algorithm, MCMC sampling is used to sample from the discrete latent space, otherwise the architectures are identical.}
    \label{fig:qaaan}
\end{figure}

Here, we demonstrate the \gls{qaaan} algorithm, Figure ~\ref{fig:qaaan}, a hybrid quantum-assited \gls{gan} in which a \gls{bm} trains, using samples from a quantum annealer, a model of a low-dimensional feature distribution of the dataset as the prior to a generator. The model learned by the algorithm is a latent variable implicit generative model $p(x \mid z) $ and an informed prior $p(z)$, where $z$ are latent variables and $x$ are data space variables. The prior will contain useful information about the features of the data distribution and this information will not need to be learned by the generator. Put another way, the prior will be a model of the feature distribution containing the latent variable modes of the dataset.

\subsection*{Contributions}

The core contribution of this work is the development of a scalable quantum-assisted \gls{gan} which trains an implicit latent variable generative model. This algorithm fulfills the criteria for inclusion of near-term quantum annealing hardware in deep learning frameworks that can learn continuous variable datasets: Resistant to noise, small number of variables, in a hybrid architecture. Additionally in this work we explore different topologies for the latent space model. The purpose of the work is to

\begin{itemize}
    \item compare different topologies to appropriately choose a graphical model, restricted by the connectivity of the quantum hardware, to integrate with the deep learning framework,
    \item design a framework for using sampling from a quantum annealer in generative adversarial networks, which may lead to architectures that encourage convergence and decrease mode collapse.
\end{itemize}

\subsection*{Outline}

First, there is a short section on the background of \gls{gan}s, quantum annealing and Boltzmann machines. In  Section~\ref{s:methods_a} an algorithm is developed to learn a latent variable generative model using samples from a quantum annealer to replace the canonical uniform noise input. We explore different models, specifically complete, symmetric bipartite and Chimera topologies, tested on a reduced stochastically binarized version of MNIST, for use in the latent space. In Section \ref{s:discussion} the results are detailed, including application of the \gls{qaaan} and a classical version of the algorithm to the MNIST dataset. The architectures are evaluated using the Inception Score and the Frech\'et Inception Distance. The algorithm is also implemented on the LSUN bedrooms dataset using classical sampling methods, demonstrating the scalability. 

%% file: sections/background.tex
\subsection*{Generative Adversarial Networks}

Implicit generative models are those which specify a stochastic procedure with which to generate data. In the case of a \gls{gan}, the generative network maps latent variables $\textbf{z}$ to images which are likely under the real data distribution, for example $\mathbf{x} = G(\mathbf{z})$, $G$ is the function represented by a neural network, $\mathbf{x}$ is the resulting image with $\mathbf{z} \sim q(\mathbf{z})$, and $q(\mathbf{z})$ is typically the uniform distribution between 0 and 1, $\mathcal{U}[0,1]$. 

Training a \gls{gan} can be formulated as a minimax game where the discriminator attempts to maximise the cross-entropy of a classifier that the generator is trying to minimise. The cost function of this minimax game is
\begin{align}
\begin{split}
V(D,G)=& \EX_{x \sim p(x)} [\log(D(x))] \\
&+ \EX_{z \sim q(z)} [\log(1 - D(G(z)))].
\end{split}\label{eq:minmaxoriginal}
\end{align}
$\EX_{x\sim p(x)}$ is the expectation over the distribution of the dataset, $\EX_{z \sim q(z)}$ is the expectation over the latent variable distribution and $D$ and $G$ are functions instantiated by a discriminative and generative neural network, respectively, and we are trying to find $\underset{G}{\min}\,\underset{D}{\max}\, V(D, G)$. The model learned is a latent variable generative model $P_{model}(x \mid z)$.

The first term in Equation~\ref{eq:minmaxoriginal} is the $\log$-probability of the discriminator predicting that the real data is genuine and the second the $\log$-probability of it predicting that the generated data is fake. In practice, \gls{ml} engineers will instead use a heuristic maximising the likelihood that the generator network produces data that trick the discriminator instead of minimising the probability that the discriminator label them as real. This has the effect of stronger gradients earlier in training \cite{goodfellow2014generative}.

\gls{gan}s are lauded for many reasons: The algorithm is unsupervised; the adversarial training does not require direct replication of the real dataset resulting in samples that are sharp \cite{wang2017generative}; and it is possible to perform the weight updates through efficient backpropagation and stochastic gradient descent.  There are also several known disadvantages. Primarily, the learned distribution is implicit. It is not straightforward to compute the distribution of the training set \cite{mohamed2016learning} unlike explicit, or prescribed, generative models which provide a parametric specification of the distribution specifying a $\log$-likelihood $\log P(\textbf{x})$ that some observed variable $\textbf{x}$ is from that distribution. This means that simple \gls{gan} implementations are limited to generation. 

Further, as outlined in the introduction, the training is prone to non-convergence \cite{barnett2018convergence}, and mode collapse \cite{thanh2018catastrophic}. This stability of \gls{gan} training is an issue and there are many hacks to encourage convergence, discourage mode-collapse and increase sample diversity including using spherical input space \cite{white2016sampling}, adding noise to the real and generated samples \cite{arjovsky2017wasserstein} and minibatch discrimination \cite{salimans2016improved}. We hypothesise that using an informed prior will decrease mode-collapse and encourage convergence. 

\begin{figure} [t]
    \centering
    \includegraphics[width=0.40\textwidth]{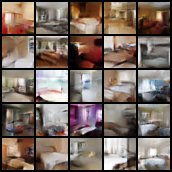}
    \caption{Bedrooms from the LSUN dataset generated with an associative adversarial network, with a fully connected latent space sampled via MCMC sampling.}
    \label{fig:lsun}
\end{figure}

\subsection*{Boltzmann Machines \& Quantum Annealing}

\begin{figure*}
\begin{minipage}{0.9\linewidth}
 \subfloat[\label{fig:fc}]{%
   \includegraphics[width=0.3\textwidth]{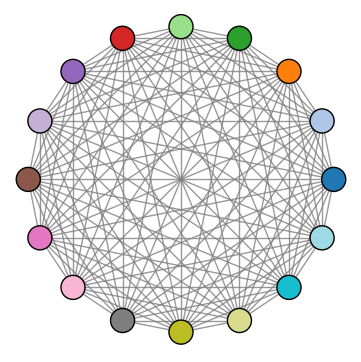}
 }
 \hfill
 \subfloat[\label{fig:sparse}]{%
   \includegraphics[width=0.3\textwidth]{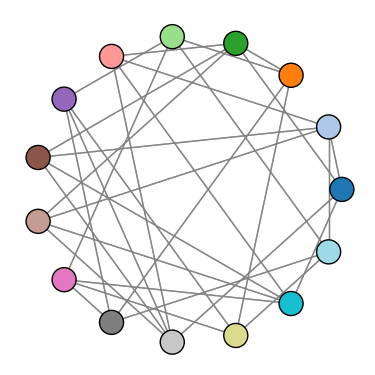}
 }
 \hfill
 \subfloat[\label{fig:restricted}]{%
   \includegraphics[width=0.3\textwidth]{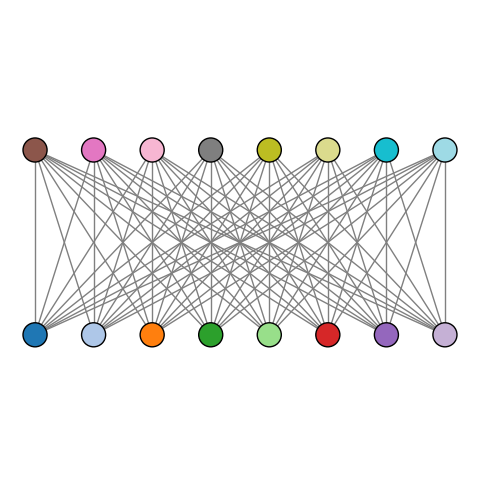}
 }
\end{minipage}
\caption{(a) Complete (b) Chimera (c) symmetric bipartite graphical models. These graphical models are embedded into the hardware and the nodes in these graphs are not necessarily representative of the embeddings.}
\end{figure*}

A \gls{bm} is a energy-based graphical model composed of stochastic nodes, with weighted connections between and biases applied to the nodes. The energy of the network corresponds to the energy function applied to the state of the system. \gls{bm}s represent multimodal and intractable distributions \cite{le2008representational}, and the internal representation of the \gls{bm}, the weights and biases, can learn a generative model of a distribution \cite{ackley1987learning}. 

A graph $\mathcal{G} = (\mathcal{V}, \mathcal{E})$ with cardinality $N$ describing a Boltzmann machine with model parameters $\boldsymbol{\lambda} = \{\boldsymbol{\omega}, \boldsymbol{b}\}$ over logical variables $\mathcal{V} = \{z_0, z_1, ... z_N\}$ connected by edges $\mathcal{E}$ has energy 
\begin{equation}
    E_{\boldsymbol{\lambda}} (\textbf{v}) = -\sum\limits_{z_i \in \mathcal{V}} b_{i} z_i - \sum\limits_{(z_i,z_j) \in \mathcal{E}} \omega_{ij} z_i z_j
    \label{eq:boltzmann_machine}
\end{equation}
where weight $\omega_{ij}$ is assigned to the edge connecting variables $z_i$ and $z_j$, bias $b_i$ is assigned to variable $z_i$ and possible states of the variables are $z_i$ $\in$ $\{-1, 1\}$ corresponding to `off' and `on', respectively. We refer to this graph as the logical graph. The distribution of the states $\textbf{z}$ is
\begin{equation}
    P(\textbf{z}) = \frac{e^{-\beta E_{\boldsymbol{\lambda}} (\textbf{z})}}{Z}
    \label{eq:boltzmann_distribution}
\end{equation}
with $\beta$ a parameter recognized by physicists as the inverse temperature in the function defining the Boltzmann distribution. 

\gls{bm} training requires sampling from the distribution represented by the energy function. For fully-connected variants it is an intractable problem to calculate the probability of the state occurring exactly \cite{koller2007graphical} and is computationally expensive to approximate. Exact inference of complete graph \gls{bm}s is generally intractable and approximate methods including Gibbs sampling are slow. Generally, applications will use deep stacked \gls{rbm} architectures, which can be efficiently trained with approximate methods.

An \gls{rbm} is a symmetric bipartite \gls{bm}. It is possible to efficiently learn the distribution of some input data spaces through approximate methods, notably contrastive divergence \cite{carreira2005contrastive}. Stacked \gls{rbm}s form a \gls{dbn} and can be greedily trained to learn the generative model of datasets with higher-level features with applications in a wide range of fields from image recognition to finance \cite{deng2014deep}. Training these types of models requires sampling from the Boltzmann distribution.

\gls{qa} has been proposed as a method for sampling from complex Boltzmann-like distributions. It is an optimisation algorithm exploiting quantum phenomena to find the ground state of a cost function. \gls{qa} has been demonstrated for a range of optimisation problems \cite{biswas2017nasa}, however, defining and detecting speedup, especially in small and noisy hardware implementations is challenging \cite{ronnow2014defining, katzgraber2014glassy}. 

\gls{qa} has been proposed and in some cases demonstrated as a sampling subroutine in \gls{ml} algorithms: A quantum Boltzmann machine \cite{amin2016quantum}; training a \gls{qvae} \cite{khoshaman2018quantum}; a quantum-assisted Helmholtz machine \cite{benedetti2018quantumqahm}; deep belief nets of stacked \gls{rbm}s \cite{adachi2015application}.

In order to achieve this, the framework outlined in Equation~\ref{eq:boltzmann_machine} can be mapped to an Ising model for a quantum system represented by the Hamiltonian
\begin{equation}
    \hat{H}_{\boldsymbol{\lambda}} = - \sum\limits_{\hat{\sigma}^z_i \in \mathcal{V}} h_i \hat{\sigma}^z_i - \sum\limits_{(\hat{\sigma}^z_i,\hat{\sigma}^z_j) \in \mathcal{E}}  J_{ij} \hat{\sigma}^z_i \hat{\sigma}^z_j.
    \label{eq:ising}
\end{equation}
where now variables $z$ have been replaced by the Pauli-$z$ operators, $\hat{\sigma}_i$, which return eigenvalues in the set $\{-1, 1\}$ when applied to the state of variable $z_i$, physically corresponding to spin-up and spin-down, respectively. Parameters $b_i$ and $\omega_{ij}$ are replaced with the Ising model parameters $h_i$ and $J_{ij}$ which are conceptually equivalent. In the hardware, these parameters are referred to as the flux bias and the coupling strength, respectively. 

The full Hamiltonian describing the dynamics of the D-Wave 2000Q, equivalent to the time-dependent transverse field Ising model, is
\begin{equation}
    \hat{H}(t) = A(t) \hat{H}_\perp + B(t) \hat{H}_{\boldsymbol{\lambda}}.
\end{equation}
The transverse field term $H_\perp$ is 
\begin{equation}
    \hat{H}_\perp = \sum\limits_{\hat{\sigma}_i^x \in V} \hat{\sigma}_i^x.
\end{equation}
$\hat{\sigma}^x$ are the Pauli-$x$ operators in the Hilbert space $\mathbb{C}^{2^N \times 2^N}$. $A(t)$ and $B(t)$ are monotonic functions defined by the total annealing time $t_\mathrm{max}$ \cite{biswas2017nasa}. Generally, at the start of an anneal, $A(0)\approx1$ and $B(0)\approx0$. $A(t)$ decreases and $B(t)$ increases monotonically with $t$ until, at the end of the anneal, $A(t_\mathrm{max})\approx0$ and $B(t_\mathrm{max}) \approx 1$. When $B(t) > 0$, the Hamiltonian contains terms that are not possible in the classical Ising model, that is those that are normalised linear combinations of classical states. 

This Hamiltonian was embedded in the D-Wave 2000Q, a system with 2048 qubits, each with degree 6. Embedding is the process of mapping the logical graph, represented by Equation~\ref{eq:ising}, to hardware. If the logical graph has degree $> 6$ or a structure that is not native to the hardware, the logical graph can still be embedded in the hardware via a 1-many mapping, that means one variable $z_i$ is represented by more than one qubit. These qubits are arranged in a `chain'(this term is used even when the set of qubits forms a small tree). A chain is formed by setting the coupling strength $J_{ij}$ between these qubits to a strong value to encourage them to take a single value by the end, but not so strong that it overwhelms the $J_{ij}$ and $h_i$ in the original problem Hamiltonian or has a detrimental effect on the dynamics. There is a sweet spot for this value. In our case, we used the maximum value available on the D-Wave 2000Q, namely $-1$. At the end of the anneal, to determine the value of a logical variable expressed as a qubit chain in the hardware a majority vote is performed: The logical variable takes the value corresponding to the state of the majority of qubits. If there is no majority a coin is flipped to determine the value of the logical variable. 

Each state found after an anneal comes from a distribution, though it is not clear what distribution the quantum annealer is sampling from. For example, in problem instances with a well defined freeze-out region, the distribution is hypothesised to follow a quantum Boltzmann distribution up to the freeze-out region where the dynamics of the system slow down and diverge \cite{amin2015searching}. If the freeze-out region is narrow then the distribution can be modelled as the classical distribution of problem Hamiltonian, $H_{\boldsymbol{\lambda}}$, at $s(t^*) = s^*$, at a higher unknown effective temperature,
\begin{equation}
    \rho = \frac{e^{-\beta \hat{H}_{\boldsymbol{\lambda}}}(t^*)}{Z} \label{eq:qbd}
\end{equation}
where $Z= \Tr[e^{-\beta \hat{H_{\boldsymbol{\lambda}}}(t^*)}]$ and we have performed matrix exponentiation. In the case where $s^* = 0$ the Hamiltonian contains no off-diagonal terms and Equation~\ref{eq:qbd} is equivalent to the classical Boltzmann distribution, Equation~\ref{eq:boltzmann_distribution}, at some temperature. $\beta$ is a dimensionless parameter which depends on the temperature of the system, the energy scale of the superconducting flux qubits and open system quantum dynamics. However, it is an open question as to when the freeze-out hypothesis holds.

\begin{figure*}
\begin{minipage}{0.9\linewidth}
    \begin{algorithm}[H]
    \caption{\label{alg:qaaan}Quantum-assisted associative adversarial network training.}
        \begin{algorithmic}[1]
        \For{epochs}
        \State Sample \textit{m} Boltzmann distribution samples from $\rho \rightarrow \boldsymbol{\phi} = \{\phi_1,\phi_2,...\phi_m\}$ using quantum annealer
        \State Sample \textit{n} examples $\boldsymbol{\phi} \rightarrow \boldsymbol{\phi}^D$ and map to logical space $\phi \mapsto \textbf{z}^D$
        \State Sample \textit{n} examples $\boldsymbol{\phi} \rightarrow \boldsymbol{\phi}^B$
        \State Sample \textit{n} examples $\boldsymbol{\phi} \rightarrow \boldsymbol{\phi}^G$ and map to logical space $\phi \mapsto \textbf{z}^G$
        \State Sample \textit{n} training data examples $\textbf{x} = \{x_1, x_2, ... x_n\}$
        \State Generate $\textbf{x}^D = G(\textbf{z}^D)$
        \State $\theta_D \leftarrow \theta_D - \nabla_{\theta_D} \sum\limits_{\textit{i}}^{\textit{n}} \big( \log D(x_i) + \log(D(x^D_i)) \big)$  \label{qaaan:D}
        \State Generate $\textbf{z}_f = D(\textbf{x})$
        \State Update weights of \gls{bm} via SGD with $\textbf{z}_f$ and $\boldsymbol{\phi}_B$
        \State Generate $\textbf{x}^G = G(\textbf{z}^G)$
        \State $\theta_G \leftarrow \theta_G - \nabla_{\theta_G} \sum\limits_{\textit{i}}^{\textit{n}} \big( \log(D(x^G_i)) \big)$ \label{qaaan:G}
        \EndFor
        \State \textbf{return} Network $G(z;\theta_G)$
        \end{algorithmic}
    \end{algorithm}
\end{minipage}
\caption{QAAAN training algorithm. $\rho$ represents the distribution given by the quantum annealer from sampling, therefore $\rho \rightarrow \phi$ represents sampling a set of vectors $\phi$ from distribution $\rho$. Steps 3 - 5 are indicative of the real-world implementation of these devices. In order to reduce sampling time we sampled from the device once and used this set for different tasks: $\phi^D$ for generating samples to train the discriminator, $\phi^B$ for training the \gls{bm} and $\phi^G$ for generating samples to train the generator. Further details on mapping to the logical space for samples from the quantum annealer can be found in Section~\ref{s:methods_a}. $\textbf{x}$ is the MNIST dataset. Steps \ref{qaaan:D} and \ref{qaaan:G} are typical of \gls{gan} implementation, $G$ and $D$ are the action of the generator discriminator network, respectively.}
\end{figure*}

Other implementations of training graphical models have accounted for this instance dependent effective temperature \cite{benedetti2016estimation}, in this work to get around the problem of using the unknown effective temperature for training a probabilistic graphical model, we use a gray-box model approach proposed in \cite{benedetti2017quantum}. In this approach, full knowledge of the effective parameters, dependent on $\beta$, are not needed to perform the weight updates as long as the projection of the gradient is positive in the direction of the true gradient. The gray-box approach ties the model generated to the specific device used to train the model, though is robust to noise and is not required to estimate $\beta$ \cite{raymond2016global}, for the purposes of Equations~\ref{eq:learning1} and \ref{eq:learning2}. We find that under this approach performance remains good enough for deep learning applications.

Though we do not have full knowledge of the distribution the quantum annealer samples from, we have modelled it as a classical Boltzmann distribution at an unknown temperature. This allows us to train models without the having to estimate the temperature of the system, providing a simple approach to integrating probabilistic graphical models into deep learning frameworks. 




%% file: sections/methods.tex
In this section, the \gls{qaaan} algorithm is outlined, including a novel way to learn the feature distribution generated by the discriminator network via a \gls{bm} using sampling from a quantum annealer. The \gls{qaaan} architecture is similar to the classical Associative Adversarial Network proposed in Ref~\cite{arici2016associative}, as such the minimax game played by the \gls{qaaan} is
\begin{align}
    \begin{split}
       V(D, G, \rho) =& \EX_{x \sim p_{\mathsf{data}}(x)}[\log D(x)]     \\&+ \EX_{z \sim \rho(z)}  [\log(1 - D(G(z)))]   \\ &+ \EX_{f \sim \rho_{f}(f)} [\log \rho],
    \end{split}
    \label{eq:minmax}
\end{align}
where the aim is now to find $\underset{\mathcal{G}}{\textsf{min}}\,\underset{\rho}{\textsf{max}}\,\underset{\mathcal{D}}{\textsf{max}}\, V(D, G, \rho)$, with equivalent terms to Equation~\ref{eq:minmaxoriginal} plus an additional term to describe the optimisation of the model $\rho$, Equation~\ref{eq:qbd}. This term conceptually represents the probability that samples generated by the model $\rho$ are from the feature distribution $\rho_f$. $\rho_f$ is the feature distribution extracted from the interim layer of the discriminator. This distribution is assumed to be Boltzmann, a common technique for modelling a complex distribution.  

The algorithm used for training $\rho$, a probabilistic graphical model, is a \gls{bm}. Sampling from the quantum annealer, the D-Wave 2000Q, replaces a classical sampling subroutine in the \gls{bm}. $\rho$ is used in the latent space of the generator, Figure~\ref{fig:qaaan}, and samples from this model, also generated by the quantum annealer, replace the canonical uniform noise input to the generator network. Samples from $\rho$ are restricted to discrete values, as the measured values of qubits are $\mathbf{z}\in\{-1,+1\}$. These discrete variables $\mathbf{z}$ are reparametrised to continuous variables $\boldsymbol{\zeta}$ before being processed by the layers of the generator network, producing `generated' data. Generated and real data are then passed into the layers of the discriminator which extracts the low-dimensional feature distribution $\rho_f$. This is akin to a variational autoencoder, where an approximate posterior maps the evidence distribution to latent variables which capture features of the distribution \cite{doersch2016tutorial}. The algorithm for training the complete network is detailed in Algorithm~\ref{alg:qaaan}. 

Below, we outline the details of the \gls{bm} training in the latent space, reparametrisation of discrete variables, and the networks used in this investigation. Additionally, we detail an experiment to distinguish the performance of three different topologies of probabilistic graphical models to be used in the latent space.  

\subsection*{Latent space}
As in Figure~\ref{fig:qaaan}, samples from a intermediate layer of the discriminator network are used to train a model for the latent space of the generator network. Here, a \gls{bm} trains this model. The cost function of this \gls{bm} is the quantum relative entropy 
\begin{equation}
    S(\rho || \rho_f ) = \Tr[\rho \ln \rho] - \Tr[\rho \ln \rho_f]
    \label{eq:quantum_relative_entropy}
\end{equation}
equivalent to the classical Kullback-Leibler divergence when all off-diagonal elements of $\rho$ and $\rho_f$ are zero. This metric measures the divergence of distribution $\rho$ from $\rho_f$ where $\rho_f$ is the target feature distribution of features extracted by the discriminator network and $\rho$ is the model trained by the \gls{bm}, from Equation~\ref{eq:minmax}. Though the distributions used here are modelled classically, this framework can be extended to quantum models using the quantum relative entropy. Given this it can be shown that the updates to the weights and biases of the model are
\begin{align}
    \Delta J_{ij} &= \eta \beta [\Braket{z_i z_j}_{\rho_f} - \Braket{z_i z_j}_{\rho}]  \label{eq:learning1} \\
    \Delta h_i &= \eta \beta [\Braket{z_i}_{\rho_f} - \Braket{z_i}_\rho]. \label{eq:learning2}
\end{align}
$\eta$ is the learning rate, $\beta$ is an unknown parameter, and $\Braket{z}_{\rho}$ is the expectation value of $z$ in distribution $\rho$. $z$ are the logical variables of the graphical model and the expectation values $\Braket{z}_{\rho}$ are estimated by averaging 1000 samples from the quantum annealer. The quantum relative entropy is minimised by stochastic gradient descent.

\subsection*{Topologies}

We explored three different topologies of probabilistic graphical models, complete, symmetric bipartite and Chimera, for the latent space. Their performance on learning a model of a reduced stochastically binarized version of MNIST was compared, in both classical sampling, Figure~\ref{fig:topology_comparison_classical}, and sampling via quantum annealing, Figure~\ref{fig:topology_comparison_quantum}, cases. The complete topology is self-explanatory, Figure~\ref{fig:fc}, restricted refers to a symmetric bipartite graph, Figure~\ref{fig:restricted}, and the sparse is the graph native to the D-Wave 2000Q, or Chimera graph, where the connectivity of the model is determined by the available connections on the hardware, Figure~\ref{fig:sparse}. 

The models were trained by minimising the quantum relative entropy, Equation~\ref{eq:quantum_relative_entropy}, and evaluated with the $L_1\mbox{-}\mathrm{norm}$, 
\begin{equation}
    L_1\mhyphen\mathrm{norm} = \sum_{z_i,z_j \in \mathcal{V}} \Braket{z_iz_j}_{\rho_f} - \Braket{z_iz_j}_\rho.
\end{equation}
The algorithm did not include temperature estimation, or methods to adjust intra-chain coupling strengths for the embedding, as in \cite{benedetti2016estimation} and \cite{benedetti2017quantum}, respectively. The method used here makes a comparison between the different topologies, though for best performance one would want to account for the embedding and adjust algorithm parameters, such as the learning rate, to each topology. 

In addition to these requirements, there are several non-functioning, `dead', qubits and couplers in the hardware. These qubits or couplers were removed in all embeddings, which had a negligible effect on the final performance. The complete topology embedding was found using a heuristic embedder. A better choice would be a deterministic embedder, resulting in shorter chain lengths, though when adjusting for the dead qubits the symmetries are broken and the embedded graph chain length increases to be comparable to that returned by the heuristic embedder. The restricted topology was implemented using the method detailed by Adachi and Henderson \cite{adachi2015application}. The Chimera topology was implemented on a 2x2 grid of unit cells, avoiding dead qubits. Learning was run over 5 different embeddings for each topology and the results averaged. For topologies requiring chains of qubits, the couplers in the chains were set to -1. 


\begin{figure}[h]
    \centering
    \includegraphics[width=0.45\textwidth]{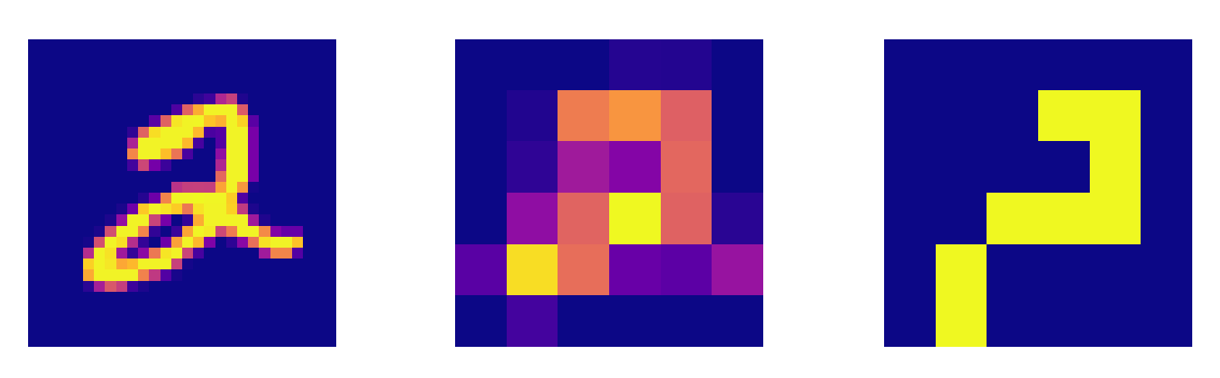}
    \caption{Left to right: 28x28 continuous, 6x6 continuous, 6x6 stochastically binarized example from the MNIST dataset.}
    \label{fig:red_bin_mnist}
\end{figure}
\subsection*{Reparametrisation}
\begin{figure}
    \centering
    \includegraphics[width=0.5\textwidth]{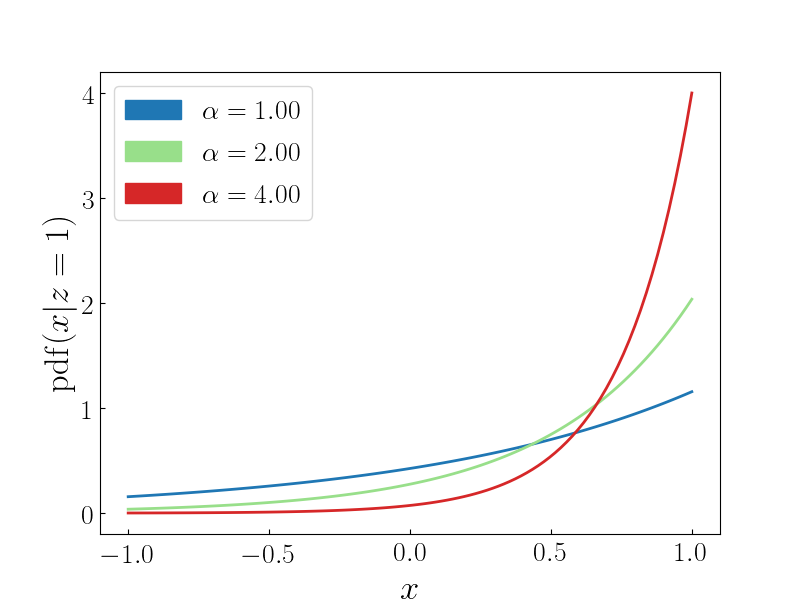}
    \caption{The probability density function, $p(x)$, for different values of $\alpha$. In this investigation $\alpha = 4$ was used, to distinguish strongly from the uniform noise case.}
    \label{fig:my_label}
\end{figure}
Samples from the latent space come from a discrete space. These variables are reparametrised to a continuous space, using standard techniques. There are many potential choices for reparametrisation functions and a simple example case is outlined below. We chose a probability density function $\mathrm{pdf}(x)$ which rises exponentially and can be scaled by parameter $\alpha$:
\begin{equation}
    \mathrm{p}(x) = \frac{\alpha \exp(-\alpha(1-x))}{1 - \exp(-2\alpha)}.
\end{equation}
The cumulative distribution function of this probability density function is 
\[   
F(z) = 
     \begin{cases}
       \int^z_{-1} p(x)\, dx &\quad -1 < z \leq 1 \\
       0 &\quad \text{otherwise,}\\
     \end{cases}
\]
and 
\begin{equation}
    \int^r_{-1} p(x)\, dx = \frac{\exp(-\alpha(1-r))-\exp(-2\alpha)}{1 - \exp(-2\alpha)}
    \label{eq:reparametrisation}
\end{equation}
Discrete samples can be reparametrised by sampling $r$ from $\mathcal{U}(-1,1]$ and inputting into Equation~\ref{eq:reparametrisation}. The value of $\alpha$ was set to $4$.

\subsection*{Networks}
The generator network consists of dense and transpose convolutional, stride 2 kernel size 4, layers with batch normalisation and ReLU activations. The output layer has a tanh activation. These components are standard deep learning techniques found in textbooks, for example \cite{Goodfellow-et-al-2016}.

The discriminator network consists of dense, convolutional layers, stride 2 kernel size 4, LeakyReLU activations. The dense layer corresponding to the feature distribution was chosen to have tanh activations in order that outputs could map to the \gls{bm}. The hidden layer representing $\rho_f$ was the fourth layer of the discriminator network with 100 nodes. When sampling the training data for the \gls{bm} from the discriminator, the variables given values from the set \{$-1, 1$\} as in the Ising model, dependent on the activation of the node being greater or less than the threshold, set at zero, respectively. 

The networks were trained with an Adam optimiser with learning rate 0.0002 and the labels were smoothed with noise. For the sparse graph latent space used in learning the MNIST dataset in Section~\ref{s:discussion}, the \gls{bm} was embedded in the D-Wave hardware using a heuristic embedder. As there is a 1-1 mapping for the sparse graph it was expressed in hardware using 100 qubits. An annealing schedule of 1$\mu s$ and a learning rate of 0.0002 were used.  The classical architecture that was compared with the \gls{qaaan} was identical other than replacing sampling via quantum annealing with MCMC sampling techniques.

%% file: sections/discussion.tex
For this work we performed several experiments. First, we compared three topologies of graphical models, trained using both classical and quantum annealing sampling methods. They were evaluated for performance by measuring the $L_1\mhyphen\mathrm{norm}$ over the course of the learning a reduced stochastically binarzied version of the MNIST dataset, Figure \ref{fig:red_bin_mnist}. Second, the \gls{qaaan} and the classical associative adversarial network described in Section~\ref{s:methods_a} were both used to generate new examples of the MNIST dataset. Their performance was evaluated used the inception score and the \gls{fid}. Finally, the classical associative adversarial network was used to generate new examples of the LSUN bedrooms dataset. 

\begin{figure}
    \centering
    \includegraphics[width=0.5\textwidth]{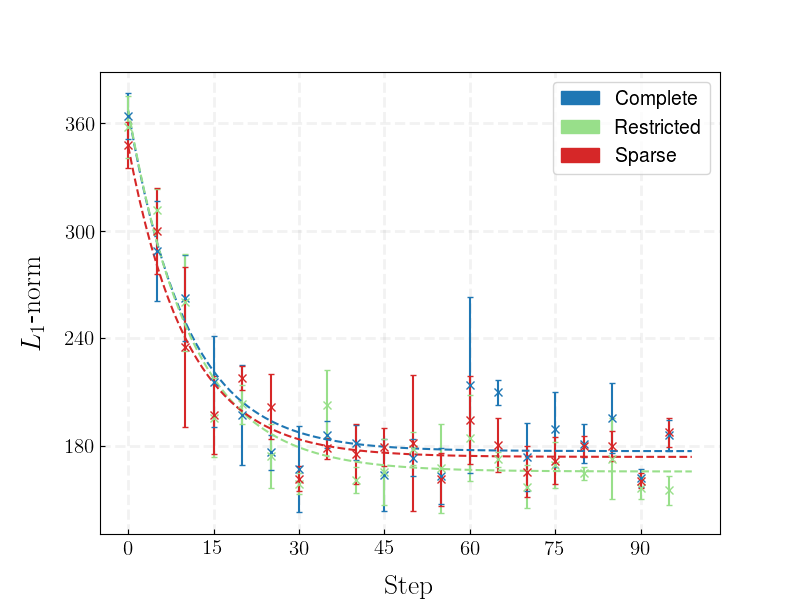}
    \caption{Comparison of the convergence of different graphical topologies trained using samples from a quantum annealers on a reduced stochastically binarized MNIST dataset. The learning rate used was 0.03. This learning rate produced the fastest learning with no loss in performance of the final model. The learning was run 5 times over different embeddings and the results averaged. The error bars describe the variance over these curves.}
    \label{fig:topology_comparison_quantum}
\end{figure}

In the experiment comparing topologies, as expected, the \gls{bm} trains a better model faster with higher connectivity, Figure~\ref{fig:topology_comparison_classical}. When trained via sampling with the quantum annealer the picture is less intuitive, Figure~\ref{fig:topology_comparison_quantum}. All topologies learned a model to the same accuracy, at similar rates. This indicates that there is a noise floor preventing the learning of a better model in the more complex graphical topologies. For the purposes of this investigation the performance of the sparse graph was demonstrated to be enough to learn an informed prior for use in the \gls{qaaan} algorithm. 

\begin{figure}[h!]
 \subfloat[\label{fig:mnist_classical}]{%
   \includegraphics[width=0.35\textwidth]{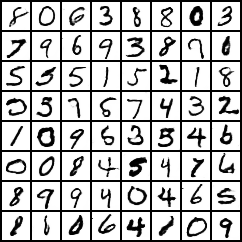}
 }
 \hfill
 \subfloat[\label{fig:mnist_quantum}]{%
   \includegraphics[width=0.35\textwidth]{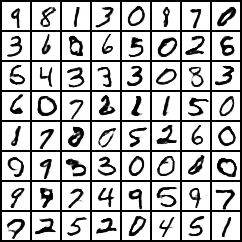}
 }
 \caption{Example MNIST characters generated by (a) classical and (b) quantum-assisted associative adversarial network architectures, with sparse topology latent spaces.}
 \label{fig:digits}
\end{figure}

Second, for the classical associative adversarial network, all topologies were implemented, and the quantum-assisted algorithm was implemented with a sparse topology latent space. The generated images for sparse topology latent spaces are shown for both classical and quantum versions in Figures~\ref{fig:mnist_classical} and \ref{fig:mnist_quantum}.

We evaluated classical and quantum-assisted versions of the associative adversarial network with sparse latent spaces via two metrics, the inception score and the \gls{fid}. Both metrics required an inception network, a network trained to classify images from the MNIST dataset, which was trained to an accuracy of $\sim95\%$. The Inception Score, Equation~\ref{eq:inception_score}, attempts to quantify realism of images generated by a model. For a given image, $p(y | x)$ should be dominated by one value of $y$, indicating a high probability that an image is representative of a class. Secondly, over the whole set there should be a uniform distribution of classes, indicating diversity of the distribution. This is expressed
\begin{equation}
    \mathrm{IS} = \exp( \EX_{x\sim\rho_\mathrm{D}} \mathrm{D}_\mathrm{KL}(p(y | x) || p(y))).
    \label{eq:inception_score}
\end{equation}
The first criterion is satisfied by requiring that image-wise class distributions should have low entropy. The second criterion implies that the entropy of the overall distribution should be high. The method is to calculate the KL distance between these two distributions: A high value indicates that both the $p(y|x)$ is distributed over one class and $p(y)$ is distributed over many classes. When averaged over all samples this score gives a good indication of the performance of the network. The inception score of the classical and quantum-assisted versions were $\sim5.7$ and $\sim5.6$, respectively.

The \gls{fid} measures the similarity between features extracted by an inception network from the dataset $X$ and the generated data $G$. The distribution of the features are modelled as a mutlivariate Gaussian. Lower FID values mean the features extracted from the generated images are closer those for the real images. In Equation~\ref{eq:frechet_inception_distance}, $\mu$ are the means of the activations of an interim layer of the inception network and $\Sigma$ are the covariance matrices of these activations. The classical and quantum-assisted algorithms scored $\sim29$ and $\sim23$, respectively.
\begin{equation}
    \mathrm{FID}(X, G) = ||\mu_X - \mu_G||^2_2 + \Tr(\Sigma_X + \Sigma_G - 2\sqrt{\Sigma_X \Sigma_G})
    \label{eq:frechet_inception_distance}
\end{equation}
The classical implementation was also used to generate images mimicking the LSUN bedrooms dataset, Figure~\ref{fig:lsun}. This final experiment was only performed as a demonstration of scalability, and no metrics were used to evaluate performance. 

\subsection*{Discussion}

\begin{figure}
    \centering
    \includegraphics[width=0.5\textwidth]{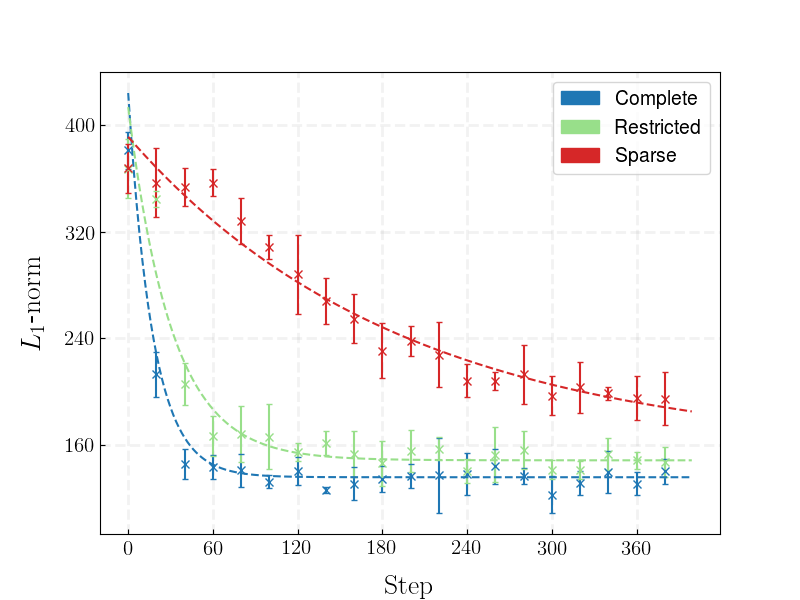}
    \caption{Comparison of different graphical topologies trained using MCMC sampling on a reduced stochastically binarized MNIST dataset. The learning rate used was 0.001. This learning rate was chosen such that the training was stable for each topology, we found that the error diverged for certain topologies at other learning rates. The learning was run 5 times and the results averaged. The error bars decribe the variance over these curves.}
    \label{fig:topology_comparison_classical}
\end{figure}

Though it is trivial to demonstrate a correlation between the connectivity of a graphical model and the quality of the learned model, Figure~\ref{fig:topology_comparison_classical}, it is not immediately clear that the benefits of increasing the complexity of the latent space can be detected easily in deep learning frameworks, such as the quantum-assisted Helmholtz machine \cite{benedetti2018quantumqahm} and those looking to exploit quantum models \cite{khoshaman2018quantum}. The effect of the complexity of the latent space model on the quality of the final latent variable generative model was not apparent in our investigations. Deep learning frameworks looking to exploit quantum hardware supported training in the latent spaces need to truly benefit from this application, and not iron out any potential gains with backpropagation. For example, if exploiting a quantum model gives improved performance on some small test problem, it is an open question as to whether this improvement will be detected when integrated into a deep learning framework, such as the architecture presented here.  

Here, given the nature of the demonstration and a desire to avoid chaining we use a sparse connectivity model. Avoiding chaining allows for larger models to be embedded into near-term quantum hardware. Given the $O(n^2)$ scaling of qubits to logical variables for a complete logical graph \cite{choi2011minor}, future applications of sampling via quantum annealing will likely exploit restricted graphical models. Though the size of near-term quantum annealers has followed Moore's law trajectory, doubling in size every two years, it is not clear what size of probabilistic graphical models will find mainstream usage in machine learning applications and exploring the uses of different models will be an important theme of research as these devices grow in size. 

There are two takeaways from the results presented here. Though these values are not comparable to state-of-the-art \gls{gan} architectures and are on a simple MNIST implementation, they serve the purpose of highlighting that the inclusion of a near-term quantum device is not detrimental to the performance of this algorithm. Secondly, we have demonstrated the framework on the larger, more complex, dataset LSUN bedrooms, Figure~\ref{fig:lsun}. This indicates that the algorithm can be scaled. 


%% file: sections/conclusions.tex
\subsection*{Summary}

In this work. we have presented a novel and scalable quantum-assisted algorithm, based on a \gls{gan} framework, which can learn a implicit latent variable generative model of complex datasets. 

This work is a step in the development of algorithms that may use quantum phenomena to improve the learning generative models of datasets. This algorithm fulfills the requirements of the three areas outlined by Perdomo-Ortiz \textit{et al} \cite{perdomo2017opportunities}: Generative problems, data where quantum correlations may be beneficial, and hybrid. This implementation also allows for use of sparse topologies, removing the need for chaining, requires a relatively small number of variables (allowing for near-term quantum hardware to be applied) and is resistant to noise.


Though the key motivation of this work is to demonstrate a functional deep learning framework integrating near-term quantum hardware in the learning process, it builds on classical work by Tarik Arici and Asli Celikyilmaz \cite{arici2016associative} exploring the effect of learning the feature space and using this distribution as the input to the generator. No claims are made here on the improvements that can be made classically, though it is possible that further research into the associative adversarial architecture will yield improvements to \gls{gan} design. 

In summary, we have successfully demonstrated a quantum-assisted \gls{gan} capable of learning a model of a complex dataset such as LSUN, and compared performance of different topologies. 

\subsection*{Further Work}

There are many avenues to use quantum annealing for sampling in machine learning, topologies and \gls{gan} research. Here, we have outlined a framework that works on simple (MNIST) and more complex (LSUN) datasets. We highlight several areas of interest that build on this work. 

The first is an investigation into how the inclusion of quantum hardware into models such as this can be detected. There are two potential improvements to the model: Quantum terms improve the model of the data distribution; or graphical models, which are classically intractable to learn for example fully connected, integrated into the latent spaces, may improve the latent variable generative model learned. Before investing extensive time and research into integrating quantum models into latent spaces it will be important to note that these improvements are reflected in the overall model of the dataset. That is, that backpropagation does not erase any latent space performance gains. 

There are still outstanding questions as to the distribution the quantum annealer samples. The pause and reverse anneal features on the D-Wave 2000Q gives greater control over the distribution output by the quantum annealer, and can be used to explore the relationship between the quantum nature of that distribution and the quality of the model trained by a quantum Boltzmann machine \cite{marshall2018power}. It is also not clear what distribution is the `best' for learning a model of a distribution. It could be that efforts to decrease the operating temperature of a quantum annealer to boost performance in optimisation problems will lead to decreased performance in \gls{ml} applications, as the diversity of states in a distribution decreases and probabilities accumulate at a few low energy states. There are interesting open questions as to the optimal effective temperature of a quantum annealer for \gls{ml} applications. This question fits within a broad are for research in \gls{ml} asking which distributions are most useful for \gls{ml} and why. 

For this simple implementation, the quantum sampling sparse graph performance is comparable to the complete and restricted topologies. Though in optimised implementations we expect divergent performance, the sparse graph serves the purpose of demonstrating the \gls{qaaan} architecture. Additionally, we have highlighted sparse classical graphical models for use in the architecture demonstrated on LSUN bedrooms. Though they have reduced expressive power there are many more applications for current quantum hardware; for example a fully connected graphical model would require in excess of 2048 qubits (the number available on the D-Wave 2000Q) to learn a model of a standard MNIST dataset, not to mention the detrimental effect of the extensive chains. A sparse D-Wave 2000Q native graph (Chimera) conversely would only use 784 qubits. This is a stark example of how sparse models might be used in lieu of models with higher connectivity. Investigations finding the optimal balance between the complexity of a model, resulting overhead required by embedding, and the affect on both on performance are needed to understand how future quantum annealers might be used for applications in \gls{ml}.

\subsection*{Acknowledgements}
We would like to thank Marcello Benedetti for conversations full of his expertise and good humour. We would also like to thank Thomas Vandal, Rama Nemani, Andrew Michaelis, Subodh Kalia and Salvatore Mandra for useful discussions and comments. 

We are grateful for support from NASA Ames Research Center, and from the NASA Earth Science Technology Office (ESTO), the NASA Advanced Exploration systems (AES) program, and the NASA Transformative Aeronautic Concepts Program (TACP). We also appreciate support from the AFRL Information Directorate under grant F4HBKC4162G001 and the Office of the Director of National Intelligence (ODNI) and the Intelligence Advanced Research Projects Activity (IARPA), via IAA 145483. The views and conclusions contained herein are those of the authors and should not be interpreted as necessarily representing the official policies or endorsements, either expressed or implied, of ODNI, IARPA, AFRL, or the U.S. Government. The U.S. Government is authorized to reproduce and distribute reprints for Governmental purpose notwithstanding any copyright annotation thereon.